\documentclass[runningheads]{llncs}
\usepackage[T1]{fontenc}
\usepackage{multirow} 
\usepackage{graphicx}
\usepackage{array}
\usepackage{graphicx}

\begin{document}
\title{Adapting and Evaluating Multimodal Large Language Models for Adolescent Idiopathic Scoliosis Self-Management: A Divide and Conquer Framework}
\titlerunning{Divide and Conquer Framework for MLLMs in AIS Self-Management}
%
%

\author{Zhaolong Wu \and Pu Luo \and Nan Meng\and Jason Pui Yin Cheung \and Teng Zhang} 
\authorrunning{Z. Wu et al.}
\institute{Department of Orthopaedics and Traumatology, The University of Hong Kong, Hong Kong SAR, China \
\email{\{wuzl01,luopudent\}@connect.hku.hk, \{nanmeng,cheungjp,tgzhang\}@hku.hk}}

%
\maketitle              
\begin{abstract}
This study presents the first comprehensive evaluation of Multimodal Large Language Models (MLLMs) for Adolescent Idiopathic Scoliosis (AIS) self-management. We constructed a database of approximately 3,000 anteroposterior X-rays with diagnostic texts and evaluated five MLLMs through a `Divide and Conquer' framework consisting of a visual question-answering task, a domain knowledge assessment task, and a patient education counseling assessment task. Our investigation revealed limitations of MLLMs' ability in interpreting complex spinal radiographs and comprehending AIS care knowledge. To address these, we pioneered enhancing MLLMs with spinal keypoint prompting and compiled an AIS knowledge base for retrieval augmented generation (RAG), respectively. Results showed varying effectiveness of visual prompting across different architectures, while RAG substantially improved models' performances on the knowledge assessment task. Our findings indicate current MLLMs are far from capable in realizing personalized assistant in AIS care. The greatest challenge lies in their abilities to obtain accurate detections of spinal deformity locations (best accuracy: 0.55) and directions (best accuracy: 0.13).

\end{abstract}
\section{Introduction}
Adolescent Idiopathic Scoliosis (AIS) is the most common spinal deformity in pediatric populations, affecting up to 4.8\% of adolescents\cite{fong2015population,de2012pathogenesis}. This condition predominantly occurs during growth spurts between ages 11 and 14. Without timely intervention, spinal deformities may progressively worsen, leading to serious consequences such as back pain and impaired pulmonary function, reducing patients' quality of life \cite{cheung2018curve}.

While treatments, surgical-interventions and clinical management are key in AIS care, patient self-management plays an essential role in the recovery from AIS, which could have paramount effects on outcomes, quality of life and long-term well-being \cite{dufvenberg2021six,karol2016effect}. This includes exercise and physical therapy, compliance with treatment, emotion and mental health as well as monitoring and reporting \cite{marchese2023world,holt2020sticking,li2024different}.

MLLMs such as LLaVA-Med have demonstrated powerful capabilities in medical image analysis, detecting lesions, analyzing conditions, and providing medical advice \cite{li2023llava,tu2024towards,moor2023med}. Although these models excel in chest radiographs and fundus image analysis, research targeting spinal diseases, particularly AIS, remains scarce due to the limited availability of specialized spinal data. Furthermore, it is unclear whether and how well current state-of-the-art open-source / weight MLLMs are at realizing a personal assistant for AIS self-management, which requires capabilities beyond medical imaging comprehension.

To address this gap, we constructed a database of approximately 3,000 anteroposterior X-rays with corresponding diagnostic texts. We evaluated leading MLLMs through a comprehensive framework that breaks down the AIS self-management requirement into three downstream tasks: a visual spinal assessment task (the ability of x-ray analysis for disease progression), a domain knowledge assessment task (the understanding of the disease and managements), and a patient education counseling assessment task (the ability to do personalized assistant given patient's x-ray and open-ended queries).


The main contributions of this paper can be summarized as follows.

\begin{enumerate}   
    \item We conducted the first comprehensive study on using MLLMs for AIS self-management through a divide-and-conquer framework, which is aligned well with clinical practice, separating complex AIS care into specialized tasks.
     \item For adapting MLLMs for AIS care, we integrated a spinal keypoint detection model with MLLMs for improving their ability on analyzing spinal deformities for the image modality. To improve MMLMs in comprehending AIS knowledge, we compiled an AIS knowledge base and implemented a knowledge augumented generation approach.
    \item We constructed a large-scale specialized image-text database for AIS, the largest to date to the best of our knowledge, providing foundational resources for future related research.
\end{enumerate}

\begin{figure}
    \centering
    \includegraphics[width=1\linewidth]{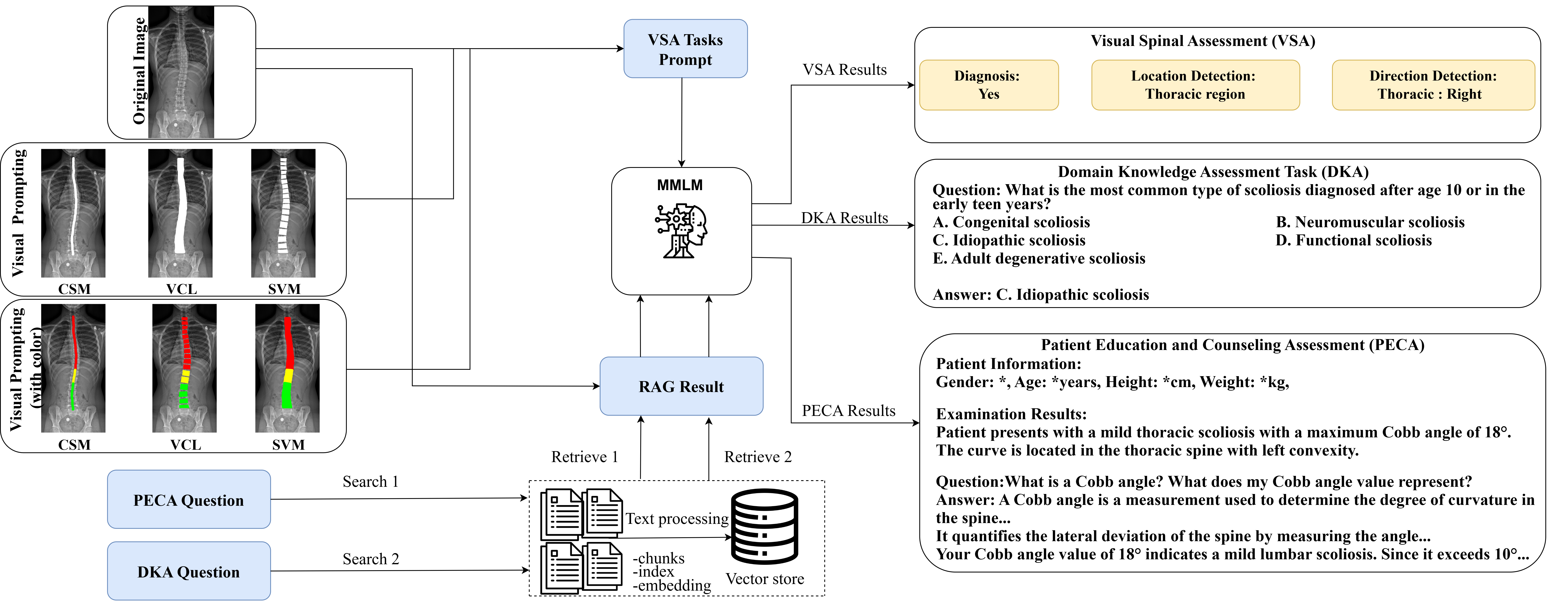}
    \caption{Divide and Conquer Framework for Adapting and Evaluating MLLMs for AIS Care.} 
    \label{fig:fig1}
\end{figure}

\section{Related Work}
Recent years have witnessed significant advances in applying MLLMs to medical tasks. These models successfully integrate various modalities to perform complex medical tasks, such as medical image-text generation, disease diagnosis, and automated report generation \cite{fan2025chestx,li2025aor,kim2024enhancing,davis2025knowledge}. However, current research exhibits a notable data bias toward certain anatomical regions.

Current MLLMs, including medical-specific models like LLaVA-Med and general models like Qwen-VL 2.5 and InternVL-3 \cite{li2023llava,bai2025qwen2,zhu2025internvl3}, have developed expertise primarily in chest radiograph analysis due to the predominance of chest X-ray datasets \cite{johnson2019mimic,irvin2019chexpert,nguyen2022vindr}. This has resulted in limited capabilities for spinal disorders like AIS, which requires precise assessment of curve patterns and specialized knowledge of classification systems and progression risk factors.

To address this gap, we introduce a novel framework that decomposes the complex AIS care process into distinct, manageable components. This "Divide and Conquer" approach allows us to systematically evaluate and adapt MLLMs for specific aspects of AIS management, from visual diagnosis to patient education. We construct a specialized spinal database and develop targeted assessment methods for each component, while also employing a keypoint detection model to supply critical anatomical landmarks and investigate what missing information leads to diagnostic failures.

\section{The `Divide and Conquer' Evaluation Framework}

Figure \ref{fig:fig1}  shows our multi-granularity clinical assessment framework for AIS analysis. We designed three comprehensive evaluation tasks to assess MLLMs' effectiveness in analyzing spinal deformities from AP spine X-rays across multiple clinical perspectives. Additionally, we integrated a spinal keypoint detection model to investigate whether diagnostic errors stem from MLLMs' inability to identify critical spinal landmarks.



\textbf{Visual Spinal Assessment (VSA)}
When diagnosing AIS in clinical practice, clinicians primarily focus on examining images to analyze spinal deformities, their location, and curve patterns. We decomposed this complex visual reasoning process into three sequential tasks of increasing granularity:

\textit{AIS Diagnosis (AD)}: This binary classification task requires MLLMs to generate a potential diagnosis based on spinal X-ray images and textual prompts, determining whether AIS is present or absent.

\textit{Spinal Deformity Location Detection (SDLD)}: This multi-class classification task focuses on localizing spinal deformities. Building on the diagnosis from AD, MLLMs must identify whether the spinal curvature occurs in the thoracic, thoracolumbar, or lumbar segments based on X-ray images and textual prompts. This localization helps determine the type of spinal deformity curve.

\textit{Spinal Deformity Direction Detection (SDDD)}: This multi-class classification task determines the direction of spinal deformity. Following AD and SDLD tasks, MLLMs must assess whether the spinal curvature is directed leftward or rightward, distinguishing between left-convex and right-convex curvatures.

\textbf{Domain Knowledge Assessment Task (DKA)}
The second major component of our framework, DKA, addresses the professional knowledge dimension of AIS care. This multiple-choice task evaluates whether MLLMs possess sufficient domain knowledge of AIS. We designed questions across six categories: basic knowledge, etiology and pathophysiology, clinical presentation and diagnosis, assessment and monitoring, treatment options, and complications and prognosis. This comprehensive assessment examines the models' understanding of AIS from fundamental concepts to diagnosis, patient management, and treatment strategies.

\textbf{Patient Education and Counseling Assessment (PECA)}
This patient-oriented question-answering task evaluates MLLMs' ability to provide accurate, accessible information to patients with varying degrees of spinal deformity. We developed 161 questions across five categories: disease explanation, treatment options, daily life management, long-term prognosis, and follow-up and monitoring. Each question was stratified by three severity levels (mild, moderate, severe) to assess how well models adapt their responses to different clinical scenarios. This task specifically measures the models' capabilities in translating complex medical knowledge into patient-appropriate explanations while maintaining clinical accuracy.

\section{Method}
\subsection{Datasets}
In this study, we collected 3,683 AP spinal X-rays with corresponding radiological reports from 3,022 patients at X Hospital between December 2019 and July 2023, with all data collection protocols approved by the Institutional Review Board (IRB). The dataset was randomly partitioned into training, validation, and testing sets following an 8:1:1 ratio, resulting in 2,946 samples for training, 368 samples for validation, and 369 samples for testing. To address the inherent class imbalance common in medical datasets, we implemented a stratified sampling approach that maintained consistent distribution of scoliosis severity across all partitions, ensuring that the proportion of each scoliosis category (normal, mild, moderate, and severe) remained constant across all sets. To minimise input image variance, all X-rays were standardized by cropping to a uniform size of 896 × 448 pixels, ensuring that the entire spinal column was captured within each image.

\subsection{Visual Prompting Strategies}
We established MLLMs zero-shot predictions as baselines and investigated how performance is affected by visual prompts provided by spine keypoint detection models. As shown in Figure \ref{fig:fig1}, we designed three visual prompting strategies: Curved Spine Midline (CSM), Vertebral Connection Line (VCL), and Segmented Vertebrae Marks (SVM). These visual prompts were designed to provide models with critical information about spinal curvature and structure. For SDLD and SDDD tasks, we differentiated thoracic, thoracolumbar, and lumbar regions using distinct colors to provide anatomical localization information.

\subsection{Retrieval-Augmented Generation (RAG) Approach}

To enhance model performance on knowledge-intensive tasks, we implemented a specialized RAG framework utilizing an AIS-specific knowledge database. This database was constructed by integrating authoritative sources including clinical practice guidelines, research publications from PubMed, and patient education resources from organizations such as the Scoliosis Research Society (SRS) \cite{kuznia2020adolescent,seifert2016adolescent,roye2020establishing,berdishevsky2016physiotherapy,dimitrijevic2024treatment}. To optimize information retrieval, we employed Gemini to generate structured knowledge graphs that capture key relationships between AIS concepts, treatments, and outcomes \cite{team2024gemini}.

\subsection{DKA and PECA Data Population}
Our evaluation datasets were designed to assess different aspects of AIS understanding. The DKA dataset consists of multiple-choice questions targeting professional medical knowledge, while the PECA dataset simulates patient-centered scenarios requiring both clinical accuracy and appropriate communication. To ensure comprehensive coverage, we developed questions spanning diagnosis, treatment options, management strategies, and daily living accommodations for patients with varying degrees of spinal deformity. A rigorous quality control process involving two junior doctors and verification by a senior physician was implemented to ensure clinical relevance, accuracy, and appropriate difficulty levels across all questions.

\subsection{Evaluation}

For the VSA task, we addressed dataset class imbalance by employing a comprehensive evaluation approach combining F1 score, AUC, and accuracy metrics. This multi-metric strategy ensures balanced assessment of model performance across both majority and minority classes. For the DKA task, we used accuracy as our primary performance metric, as it directly measures the models' proficiency in selecting correct answers within the multiple-choice format. In evaluating the PECA task, we implemented a structured human evaluation protocol wherein three junior physicians independently assessed model responses using a five-dimensional framework on a 5-point Likert scale. The detailed assessment criteria, outlined in Table 1, were specifically designed to evaluate both clinical accuracy and communication effectiveness, providing comprehensive insight into models' capabilities in addressing patient concerns across varying degrees of spinal deformity. 

\section{Experiments}

\subsection{Experimental Setup}

All experiments were conducted using 4 NVIDIA GeForce RTX 3090 GPUs. Table \ref{tab:model_description} presents the models used for comparison. We primarily utilized open-source general-domain models, selecting five multimodal large language models from four different companies, with parameter sizes ranging from 4.2B to 14B. To ensure experimental reproducibility and consistency, we standardized all MLLM configurations with a temperature parameter of 0, bfloat16 quantization, and disabled flash attention. For the PECA Task, we set the maximum response length to 300 tokens. These settings minimized model generation randomness and ensured result reliability. For keypoint detection, we employed SpineHRNet+, a model specifically trained on spine data based on HRNet and UNet architectures \cite{meng2022artificial}.
\begin{table}
\centering
\caption{Multimodal Large Language Models Description.}
\label{tab:model_description}
\resizebox{\textwidth}{!}{%
\begin{tabular}{|l|l|l|l|l|l|}
\hline
\textbf{Model} & \textbf{Vision Encoder} & \textbf{Backbone LLM} & \textbf{Connector} & \textbf{Release dates}\\
\hline
Qwen2.5-VL-7B  &  Redesigned ViT   & Qwen2.5-7B  & MLP&2025.01  \\
InternVL3-8B  &  InternViT-300M-448px-V2\_5   & Qwen2.5-7B  & MLP&2024.04   \\
InternVL3-14B  &  InternViT-300M-448px-V2\_5   & Qwen2.5-14B  & MLP& 2025.04\\
Llama 3.2-Vision  &  CLIP ViT-H/14   & Llama 3.1-8B  & MLP&2024.07  \\
Phi-3-Vision  &  CLIP ViT-L/14   & Phi-3 Mini  & Projection  &2024.05  \\
\hline
\end{tabular}
}
\end{table}

\subsection{Results and Discussion}

\begin{table}
\centering
\caption{Visual Spinal Assessment Results. Thor. = Thoracic, TL = Thoracolumbar, Lum. = Lumbar, OA = Overall accuracy. Baseline shows results without visual prompts. "Color" indicates region-specific color encoding in visual prompts.}
\label{tab:QSA_results}
\resizebox{\textwidth}{!}{%
\begin{tabular}{|l|l|cc|c@{\hspace{3mm}}c@{\hspace{3mm}}|c@{\hspace{3mm}}c@{\hspace{3mm}}|cccc|}
\hline
\multirow{3}{*}{\textbf{Model}} & \multirow{3}{*}{\textbf{Method}} & \multicolumn{2}{c|}{\textbf{Task 1: AIS}} & \multicolumn{4}{c|}{\textbf{Task 2: SDLD (OA-Acc, Region-F1,AUC)}} & \multicolumn{4}{c|}{\textbf{Task 3: SDDD(OA-Acc, Region-Acc)}} \\
\cline{3-12}
 & & \multirow{2}{*}{\textbf{F1}} & \multirow{2}{*}{\textbf{AUC}} & \multicolumn{2}{c|}{\textbf{No Color}} & \multicolumn{2}{c|}{\textbf{With Color}} & \multicolumn{2}{c|}{\textbf{No Color}} & \multicolumn{2}{c|}{\textbf{With Color}} \\
\cline{5-12}
 & & & & \textbf{OA} & \textbf{Thor./TL/Lum.} & \textbf{OA} & \textbf{Thor./TL/Lum.} & \textbf{OA} & \textbf{Thor./TL/Lum.} & \textbf{OA} & \textbf{Thor./TL/Lum.} \\
\hline
\multirow{4}{*}{Qwen2.5-VL-7B} & Baseline & 0.83 & 0.74 & 0.15 & 0.19,0.52/0.22,0.52/0.51,0.58 & - & - & 0.09 & 0.37/0.45/0.43 & - & - \\
 & CSM & 0.93 & 0.79 & 0.33 & 0.68,0.64/0.25,0.57/0.61,0.62 & 0.31 & 0.79,0.69/0.17,0.47/0.71,0.57 & 0.09 & 0.49/0.51/0.36 & 0.13 & 0.58/0.53/0.30 \\
 & VCL & 0.96 & 0.71 & 0.43 & 0.81,0.67/0.15,0.49/0.71,0.58 & 0.44 & 0.81,0.77/0.23,0.55/0.80,0.64 & 0.08 & 0.58/0.66/0.24 & 0.08 & 0.64/0.64/0.22 \\
 & SVM & 0.92 & 0.52 & 0.35 & 0.81,0.55/0.19,0.51/0.75,0.48 & 0.40 & 0.78,0.57/0.11,0.47/0.77,0.54 & 0.03 & 0.52/0.67/0.17 & 0.12 & 0.51/0.69/0.32 \\
\hline
\multirow{4}{*}{InternVL3-8B} & Baseline & 0.94 & 0.50 & 0.10 & 0.38,0.59/0.24,0.54/0.20,0.50 & - & - & 0.03 & 0.42/0.22/0.32 & - & - \\
 & CSM & 0.94 & 0.50 & 0.07 & 0.82,0.56/0.23,0.51/0.02,0.50 & 0.06 & 0.80,0.50/0.23,0.50/0.00,0.50 & 0.01 & 0.53/0.07/0.36 & 0.00 & 0.49/0.05/0.36 \\
 & VCL & 0.94 & 0.50 & 0.07 & 0.82,0.56/0.23,0.51/0.02,0.50 & 0.05 & 0.81,0.53/0.23,0.50/0.00,0.50 & 0.01 & 0.53/0.07/0.36 & 0.00 & 0.51/0.05/0.36 \\
 & SVM & 0.94 & 0.50 & 0.07 & 0.83,0.62/0.23,0.50/0.02,0.50 & 0.06 & 0.80,0.50/0.23,0.50/0.00,0.50 & 0.01 & 0.56/0.05/0.36 & 0.00 & 0.49/0.05/0.36 \\
\hline
\multirow{4}{*}{InternVL3-14B} & Baseline & 0.60 & 0.57 & 0.26 & 0.57,0.56/0.00,0.50/0.48,0.53 & - & - & 0.10 & 0.46/0.87/0.30 & - & - \\
 & CSM & 0.95 & 0.65 & 0.17 & 0.82,0.55/0.00,0.50/0.02,0.50 & 0.28 & 0.79,0.77/0.00,0.50/0.43,0.54 & 0.11 & 0.52/0.87/0.36 & 0.16 & 0.61/0.87/0.34 \\
 & VCL & 0.96 & 0.83 & 0.21 & 0.84,0.63/0.00,0.50/0.02,0.50 & 0.32 & 0.83,0.80/0.04,0.51/0.49,0.53 & 0.14 & 0.57/0.87/0.36 & 0.21 & 0.65/0.87/0.35 \\
 & SVM & 0.92 & 0.84 & 0.22 & 0.83,0.67/0.00,0.50/0.25,0.52 & 0.50 & 0.82,0.74/0.00,0.50/0.79,0.66 & 0.16 & 0.60/0.87/0.37 & 0.38 & 0.62/0.87/0.57 \\
\hline
\multirow{4}{*}{Llama 3.2-Visio} & Baseline & 0.94 & 0.50 & 0.05 & 0.81,0.53/0.24,0.55/0.80,0.58 & - & - & 0.02 & 0.50/0.17/0.17 & - & - \\
 & CSM & 0.94 & 0.50 & 0.03 & 0.80,0.50/0.23,0.52/0.74,0.54 & 0.01 & 0.81,0.52/0.23,0.50/0.79,0.51 & 0.01 & 0.49/0.09/0.17 & 0.00 & 0.50/0.05/0.10 \\
 & VCL & 0.80 & 0.77 & 0.10 & 0.75,0.67/0.23,0.54/0.70,0.63 & 0.10 & 0.74,0.66/0.23,0.54/0.73,0.64 & 0.09 & 0.57/0.38/0.27 & 0.09 & 0.56/0.38/0.27 \\
 & SVM & 0.94 & 0.50 & 0.06 & 0.80,0.50/0.23,0.50/0.34,0.44 & 0.01 & 0.80,0.50/0.23,0.50/0.78,0.50 & 0.00 & 0.49/0.05/0.26 & 0.00 & 0.49/0.05/0.09 \\
\hline
\multirow{4}{*}{Phi-3-Vision} & Baseline & 0.84 & 0.57 & 0.11 & 0.00,0.50/0.00,0.50/0.00,0.50 & - & - & 0.11 & 0.33/0.87/0.36 & - & - \\
 & CSM & 0.94 & 0.50 & 0.11 & 0.00,0.50/0.00,0.50/0.00,0.50 & 0.11 & 0.00,0.50/0.00,0.50/0.00,0.50 & 0.11 & 0.33/0.87/0.36 & 0.11 & 0.33/0.87/0.36 \\
 & VCL & 0.67 & 0.74 & 0.11 & 0.00,0.50/0.00,0.50/0.00,0.50 & 0.11 & 0.00,0.50/0.00,0.50/0.00,0.50 & 0.11 & 0.33/0.87/0.36 & 0.11 & 0.33/0.87/0.36 \\
 & SVM & 0.94 & 0.60 & 0.11 & 0.00,0.50/0.00,0.50/0.00,0.50 & 0.11 & 0.00,0.50/0.00,0.50/0.00,0.50 & 0.11 & 0.33/0.87/0.36 & 0.11 & 0.33/0.87/0.36 \\
\hline
\end{tabular}
}
\end{table}

\subsubsection{Impact of Different Visual Prompts on AIS Diagnosis.}
Table \ref{tab:QSA_results} shows significant variations in the performance of AIS diagnosis between models with different visual prompts. Structured visual cues generally improved diagnostic accuracy, with VCL achieving high F1 scores for Qwen2.5 (0.96) and InternVL3-14B (0.96). CSM boosted InternVL3-14B's F1 score from 0.60 to 0.95. Notably, InternVL3-8B and most Llama-3.2 configurations showed AUC=0.50, indicating they function as constant positive predictors rather than discriminative classifiers. Only VCL enabled meaningful discrimination in Llama-3.2 (AUC=0.77), while Phi-3.5 showed improved discrimination with VCL (AUC=0.74) and SVM (0.60). These findings demonstrate that visual prompts' effectiveness varies by model architecture, with certain prompting strategies uniquely enabling discriminative capabilities absent in baseline conditions.

\subsubsection{Impact of Different Visual Prompts on SDLD.}
Qwen2.5's accuracy improved significantly from a baseline of 0.15 to 0.43 with VCL prompting, while InternVL3-14B initially declined from 0.26 to 0.17 (CSM) and 0.22 (SVM). Color encoding produced mixed overall effects but dramatically enhanced InternVL3-14B with SVM (0.22→0.50). Regional performance was significantly impacted by color enhancement: for thoracic detection, Qwen2.5 maintained strong F1 scores (CSM: 0.68→0.79, VCL: 0.81→0.81, SVM: 0.81→0.78), while InternVL3-14B showed impressive gains (CSM: 0.52→0.79, SVM: 0.63→0.82). Thoracolumbar detection remained challenging with modest improvements for Qwen2.5 (F1 scores 0.55-0.67), while lumbar region detection showed no significant improvement across models with color enhancement. Results indicate that despite interventions successfully improving some models, others like InternVL3-8B, Phi-3.5, and certain Llama-3.2 configurations remained unimproved (OA ~0.10-0.11, F1=0, AUC=0.50), likely due to insufficient domain knowledge in these models. This highlights the importance of foundational model capabilities.

\subsubsection{Impact of Different Visual Prompts on SDDD.}
The best overall performance without color comes from InternVL3-14B+SVM (0.16), while with color enhancement, this same configuration dramatically improves to 0.38. Qwen2.5 shows moderate performance, with its best configuration being VCL without color for regional detection (0.58/0.66/0.24) but lower overall accuracy (0.08), suggesting it can identify individual region directions but struggles to integrate these into correct overall bending states. Color enhancement generally improves Qwen2.5's performance, particularly with CSM prompting (0.10→0.13). In stark contrast, several models demonstrate consistently poor performance regardless of prompting or color enhancement: InternVL3-8B achieves near-zero overall accuracy with color (0.0000 across all prompting strategies) and very poor thoracolumbar detection (0.04-0.06); Llama3.2 with SVM prompting similarly achieves 0.00 overall accuracy with both color versions; and multiple configurations show minimal response to different prompting strategies. Regional analysis reveals an important pattern: the seemingly high thoracolumbar accuracy (0.87) for InternVL3-14B and Phi-3.5 is misleading, as Task 2 results indicate these models are consistently outputting negative results rather than actually detecting deformities. 

\subsubsection{Enhancement Models Through RAG.}
DKA and PECA results consistently demonstrate significant improvements through RAG implementation across all models. In DKA, InternVL 2.5-14B achieved the highest accuracy (0.97 with RAG), while Phi 3.5-Vision showed the largest improvement (+0.20). Similarly, for PECA, RAG substantially enhanced Medical Accuracy (+1.06 to +1.09) and Safety (+0.94 to +1.02) across all models, though Communication Clarity saw more modest gains (+0.49 to +0.68). This suggests that while RAG effectively addresses knowledge limitations in specialized domains like AIS, extensive retrieved content may occasionally impact narrative coherence. Notably, performance gaps between models narrowed with RAG implementation, with smaller models showing proportionally greater improvements. These findings confirm that retrieval augmentation effectively compensates for limited parameters in specialized medical applications, enabling smaller models to approach the performance of larger counterparts.

\begin{table}
\centering
\caption{MLLMs Performance Comparison on DKA and PECA Tasks With and Without RAG. "w/o RAG" = Without RAG, "w/ RAG" = With RAG, "Imp" = Improvement.}
\label{tab:performance_comparison}
\resizebox{\textwidth}{!}{%
\begin{tabular}{|l|ccc|ccc|ccc|ccc|ccc|ccc|}
\hline
\multirow{3}{*}{\textbf{Model}} & \multicolumn{3}{c|}{\textbf{DKA (Acc)}} & \multicolumn{15}{c|}{\textbf{PECA (Acc)}} \\
\cline{2-19}
 & \multirow{2}{*}{\textbf{w/o RAG}} & \multirow{2}{*}{\textbf{w/ RAG}} & \multirow{2}{*}{\textbf{Imp}} & \multicolumn{3}{c|}{\textbf{Medical Accuracy}} & \multicolumn{3}{c|}{\textbf{Response Completeness}} & \multicolumn{3}{c|}{\textbf{Communication Clarity}} & \multicolumn{3}{c|}{\textbf{Response Personalization}} & \multicolumn{3}{c|}{\textbf{Safety}} \\
\cline{5-19}
 & & & & \textbf{w/o RAG} & \textbf{w/ RAG} & \textbf{Imp} & \textbf{w/o RAG} & \textbf{w/ RAG} & \textbf{Imp} & \textbf{w/o RAG} & \textbf{w/ RAG} & \textbf{Imp} & \textbf{w/o RAG} & \textbf{w/ RAG} & \textbf{Imp} & \textbf{w/o RAG} & \textbf{w/ RAG} & \textbf{Imp} \\
\hline
InternVL 2.5-14B & 0.82 & 0.97 & +0.16 & 2.96 & 3.95 & +0.98 & 3.20 & 3.93 & +0.73 & 3.38 & 3.88 & +0.50 & 3.13 & 3.84 & +0.71 & 2.87 & 3.88 & +1.02 \\
\hline
InternVL 2.5-8B & 0.77 & 0.91 & +0.14 & 2.88 & 3.84 & +0.96 & 3.11 & 3.95 & +0.84 & 3.30 & 3.86 & +0.56 & 3.08 & 3.75 & +0.67 & 2.87 & 3.84 & +0.97 \\
\hline
Llama 3.2-Vision & 0.77 & 0.94 & +0.17 & 2.84 & 3.90 & +1.06 & 3.15 & 3.89 & +0.74 & 3.30 & 3.80 & +0.50 & 3.20 & 3.79 & +0.59 & 2.84 & 3.78 & +0.94 \\
\hline
Phi 3.5-Vision & 0.70 & 0.90 & +0.20 & 2.80 & 3.88 & +1.08 & 2.89 & 3.88 & +0.99 & 3.20 & 3.88 & +0.68 & 2.88 & 3.67 & +0.79 & 2.84 & 3.82 & +0.98 \\
\hline
Qwen 2.5VL-7B & 0.78 & 0.94 & +0.16 & 2.86 & 3.95 & +1.09 & 3.09 & 3.91 & +0.82 & 3.27 & 3.76 & +0.49 & 2.89 & 3.72 & +0.83 & 2.88 & 3.84 & +0.96 \\
\hline
\end{tabular}
}
\end{table}

\section{Conclusion}
This research introduces a novel Divide and Conquer framework that breaks down complex AIS analysis into distinct, evaluable stages. Findings demonstrate that current MLLMs remain insufficient for implementing automated AIS patient self-management systems, despite showing promise in specialized tasks. While anatomical guidance improved quantification performance for models with adequate baseline capabilities, RAG significantly enhanced models' capabilities in specialized AIS knowledge domains, particularly overcoming knowledge limitations in patient education and domain knowledge tasks. This systematic evaluation approach provides a roadmap for targeted improvements, suggesting that as MLLMs advance in both foundational capabilities and specialized medical understanding, they will increasingly support clinical practice without yet replacing human expertise in AIS care.

\subsubsection{\ackname}This research was supported by the Health and Medical Research Fund (HMRF) [Grant No. 19200911 and 21223141] and the National Natural Science Foundation of China (NSFC) Young Scientists Fund [Grant No. 82303957]. We sincerely thank all funding agencies for their generous support.
\begin{credits}

\subsubsection{\discintname}
The authors have no competing interests to declare that are relevant to the content of this article.
\end{credits}
%
%
%
\bibliographystyle{splncs04}
\bibliography{Paper-0006}
%




\end{document}